\documentclass{article}
\usepackage{spconf,amsmath,graphicx,hyperref}
\usepackage{booktabs}
\usepackage{tabularx}
\usepackage{float}
\usepackage{xurl}
\usepackage{hyperref}
\usepackage{microtype}

\makeatletter
\let\orig@thebibliography\thebibliography
\def\thebibliography#1{%
  \orig@thebibliography{#1}%
  \setlength{\itemsep}{0.15ex plus 0.2ex minus 0.1ex}%
  \setlength{\parsep}{0pt}%
  \setlength{\parskip}{0pt}%
}
\makeatother
\title{Unit-Based Agent for Semi-Cascaded Full-Duplex Dialogue Systems}
\name{
Haoyuan Yu$^{1, 2\ast}$\thanks{*Equal contribution. $\dagger$Corresponding author.} \qquad
Yuxuan Chen$^{3\ast\dagger}$ \qquad
Minjie Cai$^{1\dagger}$ \qquad
}

\address{
$^{1}$ Hunan University \quad
$^{2}$ Gongdao Technology \quad
$^{3}$ Jilin University \quad
}

\begin{document}
%
\maketitle
\begin{abstract}
Full-duplex voice interaction is crucial for natural human computer interaction. We present a framework that decomposes complex dialogue into minimal conversational units, enabling the system to process each unit independently and predict when to transit to the next. This framework is instantiated as a semi-cascaded full-duplex dialogue system built around a multimodal large language model, supported by auxiliary modules such as voice activity detection (VAD) and text-to-speech (TTS) synthesis. The resulting system operates in a train-free, plug-and-play manner. Experiments on the HumDial dataset demonstrate the effectiveness of our framework, which ranks second among all teams on the test set of the Human-like Spoken Dialogue Systems Challenge\cite{zhao2026icassp2026humdialchallenge} (Track 2: Full-Duplex Interaction). Code is available at the GitHub repository \url{https://github.com/yu-haoyuan/fd-badcat}.
\end{abstract}
\begin{keywords}
Full-Duplex Dialogue Systems
\end{keywords}
\section{Introduction}

Real-world voice interaction remains a challenging task\cite{chang2025game}\cite{wang2025towards}. Full-duplex interaction offers a more natural communication paradigm, unlike half-duplex systems that prevent users and machines from speaking concurrently. Full-duplex interaction is implemented through two paradigms, end-to-end models\cite{arora2025chain} and cascaded pipelines\cite{chen2025turn}. Paralinguistic cues such as emotion and prosody\cite{cheng2025miku} play a crucial role in spoken dialogue. End-to-end models process user and agent speech jointly and preserve acoustic cues, but require heavy data and training. Cascaded pipelines built on TTS, automatic speech recognition(ASR), large language model (LLM), and speaker verification(SV) are easy to deploy and integrate. However, their multi-stage design loses paralinguistic cues and adds latency.

Cascaded pipelines remain attractive in practice because their pluggable control modules support barge-in, backchannels, and flexible component integration. Recent advances in multimodal large language model (MLLM) allow the models to use acoustic cues directly, making cascaded pipelines simpler.
\cite{wang2024full} uses an LLM to predict Speak/Listen state tokens that explicitly govern turn-taking. EasyTurn \cite{EasyTurn}\cite{geng2025osum} categorizes user utterances into four states: complete, incomplete, wait, and backchannel, making the model’s predictions more interpretable and accurate. FireRedChat\cite{chen2025fireredchat} presents a practical pluggable full-duplex framework, and StepFun\cite{wu2025chronological} further explores chronological thinking in end-to-end full-duplex spoken dialogue models to improve fine-grained speech handling.

In this work, we propose a new pluggable full-duplex dialogue framework which unifies turn-taking signals and user utterance states into a single decision space, enabling the MLLM to make coherent state-dependent decisions.
The framework is composed of four modular components:
(1) an audio acquisition module which employs VAD and SV to monitor the incoming speech stream;
(2) a context module which uses an asynchronous ASR to provide user transcripts;
(3) a decision module powered by a MLLM;
and (4) a speech generation module which synthesizes responses through TTS.
Relying on the MLLM’s built-in multimodal reasoning, the framework is train-free, and uses ASR transcripts only as contextual augmentation, thereby realizing a semi-cascaded design.
Experimental results show that the proposed framework achieves state-of-the-art semantic and interaction-state inference with reduced response latency on the HumDial dataset.

\begin{figure*}[t]
    \centering
    \includegraphics[width=1\textwidth]{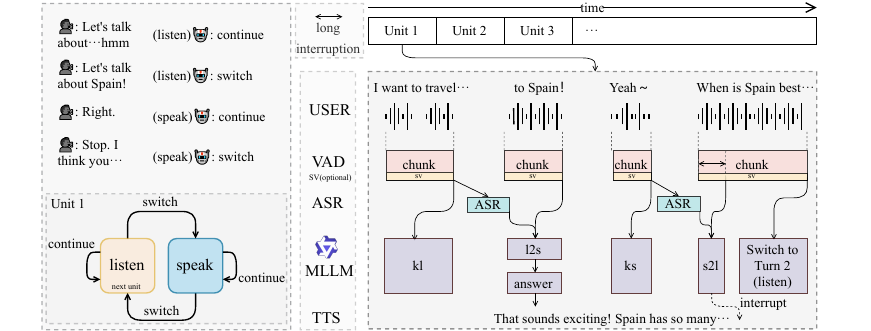}
    \caption{Overview of the unit-based dialogue process. Top-left shows example MLLM control behaviors; bottom-left shows intra-unit listen/speak transitions driven by continue/switch signals. Right shows one unit’s execution, where kl (keep listen) / l2s (listen-to-speak) correspond to continue/switch in the listen state and ks (keep speak) / s2l (speak-to-listen) correspond to continue/switch in the speak state.}
    \label{fig:result1}
\end{figure*}

\section{Method}
Our framework decomposes natural dialogue into a sequence of dialogue units.
Each unit contains two states: listen and speak.
The system always begins a unit in the listen state, and the MLLM issues one of two actions—\textit{continue} or \textit{switch}—to drive state transitions.
\textit{Continue} keeps the system in the current state, while \textit{switch} moves the system to the other state.
A switch from listen to speak occurs within the same unit, whereas a switch from speak back to listen marks the beginning of the next unit.
Through these two actions, the MLLM controls both state transition and unit progression.

To determine state transition, we simplify and reorganize the categories introduced in EasyTurn \cite{EasyTurn}. 
In the listen state, the system determines whether the user’s utterance is semantically complete, with incomplete utterances outputting a continue decision and complete ones outputting a switch decision that moves the system to the speak state, where the system generates the assistant’s spoken response.
In the speak state, the system distinguishes between user backchannels and genuine interruptions, with backchannels outputting a continue decision and interruptions outputting a switch decision that returns the system to the listen state.

The audio acquisition module continuously captures the user’s speech stream, optionally applying speaker verification to filter non-target voices, and forwards the audio to the MLLM together with a state-specific prompt.
In parallel, the audio is sent to the context module, where an asynchronous ASR generates transcripts that are cached and used in the next decision cycle as semantic augmentation.
During the speak state, the TTS module renders the assistant’s response, and any user interruption immediately terminates playback and returns the system to the listen state to begin a new unit.

\vspace{-0.7\baselineskip}
\section{experiments}
\vspace{-0.7\baselineskip}
Our system is implemented using open-source components. The audio acquisition module uses Silero VAD\footnote{\url{https://github.com/snakers4/silero-vad}}
to detect speech activity in real time, with an optional speaker-verification model CAM++\footnote{\url{https://modelscope.cn/models/damo/speech_campplus_sv_zh-cn_16k-common}}
to identify and filter target speakers. The context module 
uses Paraformer\footnote{\url{https://github.com/k2-fsa/sherpa-onnx}}, which provides ASR results as auxiliary semantic context. The multimodal decision module uses Qwen3-Omni\footnote{\url{https://github.com/QwenLM/Qwen3-Omni}}
to integrate acoustic cues and contextual information for unit-level control decisions. The speech generation module adopts IndexTTS1.5\footnote{\url{https://github.com/Ksuriuri/index-tts-vllm}}
to synthesize agent responses with streaming support.
Qwen3-Omni and IndexTTS 1.5 are vLLM-based, and the system prompts are available in our GitHub repository.
Experiments are conducted on two NVIDIA A100 GPUs using the dataset provided by the challenge organizers, with evaluation largely following Full-Duplex-Bench v1.5\cite{lin2025full} and additional metrics for full-duplex assessment.
\begin{table}[H]
  \caption{Experimental Results on the Dev and Test Set}
  \label{tab:exp-results}
  \centering
  \footnotesize

  \newcommand{\thA}{\parbox[c]{1.3cm}{\centering First\\Response\\Delay}}
  \newcommand{\thB}{\parbox[c]{1.6cm}{\centering Interruption\\Total\\Score}}
  \newcommand{\thC}{\parbox[c]{1.3cm}{\centering Rejection\\Total\\Score}}
  \newcommand{\thD}{\parbox[c]{1.1cm}{\centering Total\\Delay}}

  \begin{tabularx}{\columnwidth}{l c c c c}
    \toprule
       & \textbf{\thA}
       & \textbf{\thB}
       & \textbf{\thC}
       & \textbf{\thD} \\
    \midrule
      baseline    & 2.753s  & 80.2 & 45.6 & 2.436s \\
      \textbf{ours(Dev)}  & \textbf{1.528s} & \textbf{89.7} & \textbf{50.0} & \textbf{1.698s} \\
      \textbf{ours(Test)} & \textbf{-} & \textbf{89.7} & \textbf{57.8} & \textbf{1.632s} \\
    \bottomrule
  \end{tabularx}
\end{table}

We implement front-end and back-end interaction using WebSocket and FastAPI to simulate real-world dialogue. The experiments show that a MLLM can replace the traditional ASR-to-LLM cascade by reasoning directly over user audio, enabling the system to capture paralinguistic cues, reduce latency, and produce more informed responses.

\section{conclusion}
We present a train-free, unit-based framework for semi-cascaded full-duplex dialogue, where a MLLM directly taking audio as input controls state transition at the unit-level. Experiments show that our proposed framework achieves accurate, low-latency turn-taking, effectively replacing the traditional ASR-to-LLM cascade with a MLLM while remaining plug-and-play with existing speech components.



\bibliographystyle{IEEEbib}
\bibliography{strings,refs}

\end{document}